# Threat Classification on Deployed Optical Networks Using MIMO Digital Fiber Sensing, Wavelets, and Machine Learning

Khouloud Abdelli[(1)], Henrique Pavani[(2-3)], Christian Dorize[(2)], Sterenn Guerrier[(2)], Haïk Mardoyan[(2)], Patricia Layec[(2)], Jérémie Renaudier[(2)]

[(1)] Nokia Bell Labs, Stuttgart, Germany, khouloud.abdelli@nokia.com
[(2)] Nokia Bell Labs, Massy, France, christian.dorize@nokia-bell-labs.com
[(3)] LTCI, Télécom Paris, Palaiseau, France

**Abstract** *We demonstrate mechanical threats classification including jackhammers and excavators, leveraging wavelet transform of MIMO-DFS output data across a 57-km operational network link. Our machine learning framework incorporates transfer learning and shows 93% classification accuracy from field data, with benefits for optical network supervision. ©2024 The Author(s)*

## Introduction

Distributed Acoustic Sensing (DAS) plays a pivotal role in monitoring the integrity of optical network infrastructure, offering unparalleled sensitivity and precise localization capabilities [1,2]. Among DAS-based solutions, Multiple Input Multiple Output Digital Fiber Sensing (MIMO-DFS) emerges, offering the ability to discern various physical events along the fiber thanks to its enhanced sensitivity [3], enabling proactive maintenance and rapid response to potential threats, all without affecting data transmission [4]. In a recent study [5], we showcased the efficacy of MIMO-DFS in detecting and localizing digging events over a 57-km standard single mode fiber (SSMF) buried in sand within a deployed network.

However, harnessing DAS data to identify events such as mechanical vibrations is challenging. Traditional methods falter in handling the complexity and variability of real-world data, necessitating the integration of machine learning (ML) techniques. While some efforts have demonstrated automated event detection and localization [6-8], building resilient models requires abundant labelled dataset- a resource often scarce in field applications. This scarcity is compounded by the intricate and dynamic nature of field conditions. To overcome this obstacle, transfer learning emerges as a promising solution, harnessing knowledge from pre-existing datasets to enhance model performance despite the constraints of limited labelled data availability [9-11].

In this paper, we introduce an ML based approach that harnesses the power of transfer learning and continuous wavelet transform (CWT) to classify digging activities. By employing CWT, we transform MIMO-DFS time series data from a previous field trial [5] into scalograms, providing a 2D representation for analysis. These scalograms are then input into a pre-trained model, originally trained on extensive image dataset, serving as a feature extractor, followed by an external random forest (RF) classifier. We validate our approach using field data, demonstrating strong performance despite the constraints of a small training dataset.

## Wavelet Transform Signature from Field Data

Data used in this paper have been recorded during a field trial reported in [5], whose framework is shown in Fig. 1. The data incorporates digging activity events near a 57-km long SSMF within a deployed optical network. The utilized MIMO-DFS setup (Fig. 1-(c)) features a narrow-linewidth laser source modulated into two mutually orthogonal probing codes [3], entering the fiber under test (FUT) with 2 dBm input power (see [5] for more details). The Rayleigh backscattered signal is routed to a dual-polarization coherent receiver in self-homodyne configuration. Continuous probing of the FUT yields theoretically perfect channel estimation for each 4.1-m gauge length segment and at each 2.6-ms sequence iteration, being unaffected by polarization fading effects [3]. The estimations are used to derive the time-series of the spatial differential phase, proportional to the perturbation induced strain at each fiber segment. The standard deviation of these signals over time is used for the detection and localization of potential mechanical threats at each segment [3-4]. Buried beneath residential, urban, and desert areas, the cable experienced provoked disturbance events at 15.5 km and 29.8 km from the fiber start, as illustrated in Fig 1-(a) and 1-(b). These locations facilitate the analysis of system response to jackhammer and excavator digging events on different ground types: hard ground (15.5 km) and sandy soil for (29.8 km), each with distinct ambient noise profiles.

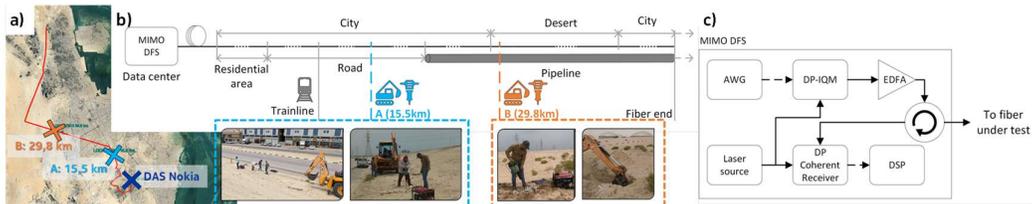

**Fig. 1:** (a) Map of the investigated fiber cable with two locations for provoked events using a jackhammer and an excavator, (b) a schematic view of the 57-km fiber link under test and field operation and (c) the MIMO-DFS system.

The analysis of events within DAS data poses challenges due to the nonstationary nature of most mechanical strains, aggravated by the ambient noise levels on the field. To overcome the limitations posed by the Power Spectrum Density (PSD) approach, we employ a CWT on the time series data obtained from the field trial, converting the data into scalograms with more discernible patterns, aiding in feature extraction. We choose a CWT with Gaussian modulated wavelets performed over 5 octaves up to the 190Hz Nyquist frequency (imposed by the probing code repetition period) and with 20 sub-bands per octave. Each disturbance source manifests a unique wavelet signature, as detailed in [5].

For each 15-second DAS recording, which encapsulates an investigated event ($C_0$: "no event", $C_1$: "jackhammer", or $C_2$: "excavator"), we perform a wavelet transform and extract 1-second windows centered around the energy peak identified in the transformed data. This process results in a dataset comprising 147 samples, with 49 samples per investigated event class. Each sample is represented by a 96x96 pixel image yielded by the wavelet signatures performed from the MIMO-DFS field trial data. Subsequently, the dataset is partitioned into 80% for training and 20% for testing. A secondary, unrelated disturbance was detected at 27.8 km from the fiber start, near an active construction site, providing an opportunity to gather an unseen test dataset to evaluate the robustness of our approach. In Fig. 2, a 15-second event at 27.8 km is shown, along with the PSD (Fig. 2-(b)), and the wavelet transform module (Fig. 2-(c)).

**ML based Digging Activity Classifier**

Fig. 3 illustrates our proposed approach for digging activity classification. Our method leverages a pre-trained deep neural network to extract features automatically. This method enables us to utilize knowledge learned from a generic image classification domain and apply it to our target task of classifying digging activities. We selected the MobileNet model [12], initially trained on the ImageNet dataset [13], due to its efficiency and accuracy compared to other pre-trained models, making it well-suited for our classification task. It is worth noting that the pre-trained MobileNet model consists of 28 layers, including convolutional layers, a fully connected layer, and a SoftMax layer with 1000 neurons.

Our proposed approach, MobileNet-RF, integrates feature extraction via MobileNet and utilizes an RF classifier. Leveraging RF's capability to handle high-dimensional data and delineate distinct decision boundaries aims to enhance classification accuracy. To fine-tune the MobileNet-RF model for our target task, we utilize a target MIMO-DFS image dataset, where time series data is transformed into scalograms using CWT as discussed previously. The features learned by MobileNet from the ImageNet dataset are transferred to this fine-tuned model, while the convolutional layers' parameters remain frozen to preserve the learned knowledge and features.

Our MobileNet-RF architecture replaces the SoftMax layer of the pre-trained MobileNet with an RF classifier connected to the last fully connected layer. The model outputs predicted class labels for different investigated event types.

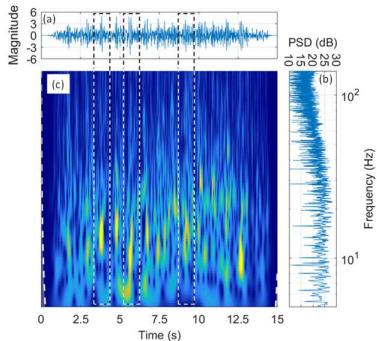

**Fig. 2:** (a) Time response, (b) PSD, and (c) scalogram of MIMO-DFS data captured at 27.8 km (construction site).

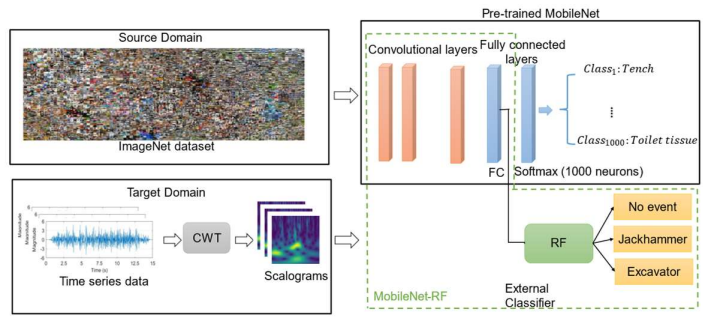

**Fig. 3:** Proposed transfer learning approach for digging activity classification.

**Results**

We adopted a 5-fold cross-validation strategy to ensure comprehensive model evaluation and validation. We initially compared the performance of our approach across various input image sizes in terms of average accuracy (Acc), F1 score (F1), and error rate percentage (Er). As depicted in Tab. 1, the results indicate that the classifier's performance improves as the input image size increases, providing more detailed information, up to (160, 160), reaching optimal performance. However, beyond this point, there is a decline in performance, as larger input sizes may introduce more noise or irrelevant information into the model, which can degrade performance.

Tab. 1: Performance Evaluation under Different Input Sizes.

| Image Input | Acc (%) | F1 (%) | Er (%) |
|---|---|---|---|
| (96, 96) | 87 | 86.7 | 12.9 |
| (128, 128) | 87.7 | 87.3 | 12.2 |
| (160, 160) | 93.2 | 93 | 6.8 |
| (192, 192) | 89 | 88.9 | 11 |

The average confusion matrix presented in Fig. 4 illustrates the effectiveness of our approach, achieving an accuracy of 93%, despite being trained on a limited dataset. The misclassification between $C_0$ and $C_1$ can be attributed to the similarity in patterns resulting from noise and the relatively low amplitude of jackhammer signals. Notably, our approach exhibits superior accuracy in classifying "excavator" events, primarily due to significantly higher signal-to-noise ratio (SNR). This higher SNR results in more meaningful scalograms, enabling finer time-frequency localization of hot spots on top of the higher signal magnitudes.

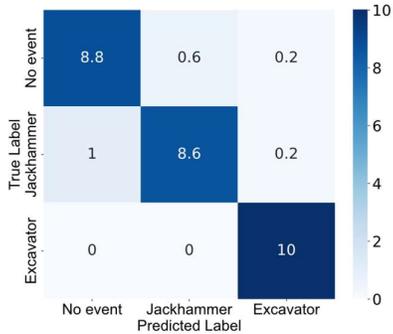

**Fig. 4:** Average confusion matrix achieved by our model.

We juxtaposed our MobileNet-RF approach against variants utilizing different classifiers: MobileNet-SoftMax and MobileNet-SVM. As shown in Tab. 2, MobileNet-RF excels, improving performance by over 2% through its adept handling of high-dimensional data and non-linear feature relationships, while also boasting lower complexity with a swift inference time of 0.2 ms compared to MobileNet-SoftMax (0.7 ms) and MobileNet-SVM (2.8 ms). Further comparisons with baseline methods, 2D-SVM and 2D-RF, which take wavelet transforms as input, reinforce our approach's superiority, showcasing more than a 16% boost across all metrics (Tab. 2). Lastly, against 1D-SVM and 1D-RF baselines, which process raw DAS time series without wavelet transformation, our approach stands out remarkably, achieving over 47.6% performance enhancement as shown in Tab. 2. This highlights the critical advantage of converting time series into scalograms, where patterns become more discernible.

Tab. 2: Performance Comparison of Classifiers Using Accuracy, F1 Score, and Error Rate Percentage Metrics.

| Method | Acc (%) | F1 (%) | Er (%) |
|---|---|---|---|
| MobileNet-RF | 93.2 | 93 | 6.8 |
| MobileNet-SVM | 91.2 | 91 | 8.7 |
| MobileNet-SoftMax | 87.7 | 87.5 | 13.8 |
| 2D-RF | 80 | 80 | 20 |
| 2D-SVM | 73.7 | 72 | 26.6 |
| 1D-RF | 63 | 63 | 36.6 |
| 1D-SVM | 60 | 59 | 40 |

To assess the robustness of our proposed approach, we employed unseen test data comprising recorded construction activities, including excavators, conducted at 27.8 km, and initiated from independent machineries. Evaluating our approach on this data yielded an accuracy of 90%, demonstrating its capability to identify excavator events accurately. Fig. 5 displays an example of model predictions with confidence intervals exceeding 0.6.

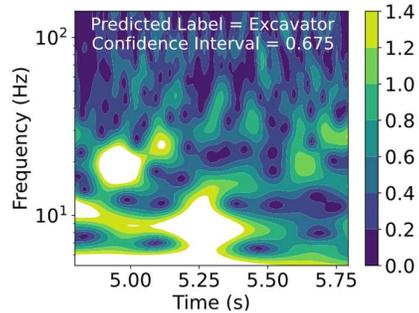

**Fig. 5:** Example of a prediction made by our approach on unseen construction work dataset.

**Conclusions**

We proposed a threat classifier leveraging transfer learning and CWT, validated using MIMO-DFS field data from a 57-km deployed network link. Results highlight the effectiveness of our method in distinguishing between different potential threats to network operation, despite being trained on a small noisy dataset. These results pave the way for improving stability and protection of optical networks using highly sensitive fiber sensing and machine learning.